# Text classification optimization algorithm based on graph neural network

Erdi Gao*
New York University
New York, USA

Haowei Yang
University of Houston
Houston, USA

Dan Sun
Washington University in St. Louis
St. Louis, USA

Haohao Xia
University of Southern California
Los Angeles, USA

Yuhan Ma
Johns Hopkins University
Baltimore, USA

Yuanjing Zhu
Duke University
Durham, USA

*Abstract*—In the field of natural language processing, text classification, as a basic task, has important research value and application prospects. Traditional text classification methods usually rely on feature representations such as the bag of words model or TF-IDF, which overlook the semantic connections between words and make it challenging to grasp the deep structural details of the text. Recently, GNNs have proven to be a valuable asset for text classification tasks, thanks to their capability to handle non-Euclidean data efficiently. However, the existing text classification methods based on GNN still face challenges such as complex graph structure construction and high cost of model training. This paper introduces a text classification optimization algorithm utilizing graph neural networks. By introducing adaptive graph construction strategy and efficient graph convolution operation, the accuracy and efficiency of text classification are effectively improved. The experimental results demonstrate that the proposed method surpasses traditional approaches and existing GNN models across multiple public datasets, highlighting its superior performance and feasibility for text classification tasks.

## I. INTRODUCTION

Text classification is a core task in Natural Language Processing [1], which is widely used in sentiment analysis[2], subject detection[3], spam filtering and other fields. Traditional text classification methods mostly rely on vector representations such as Bag-of-Words or TF-IDF, which treat text as an independent lexical set. By overlooking the semantic relationships and contextual information between words, it becomes challenging to capture the deep structural features of text. Convolutional Neural Networks [4-5] and Recurrent Neural Networks [6] have been applied to text classification tasks. Although these methods have improved classification performance to some extent, they still face limitations in capturing long-distance dependencies and processing non-Euclidean data.

Graph Neural Networks, as a kind of deep learning model that can effectively process non-Euclidean data, have made remarkable progress in many fields in recent years[7]. GNNs realizes the learning and representation of graph-structured data through the information transfer and aggregation between nodes and their neighbors, which is especially suitable for text data with complex relational structure[8]. However, the existing GNN-based text classification methods still face several challenges in practical application. On the one hand, the graph structure construction of text usually relies on word co-occurrence or syntactic dependency, which has problems of high computational complexity and unstable graph construction quality; On the other hand, the training and reasoning process of GNN model is time-consuming and requires high resources, which may not adequately address the requirements of large-scale text classification tasks.

This paper presents an optimized text classification algorithm leveraging graph neural networks. The objective is to improve the efficiency and accuracy of text classification by optimizing the graph construction strategy and the graph convolution process. Specifically, this paper uses the adaptive graph construction method to dynamically adjust the relationship between nodes and edges according to the text content, so as to build a more accurate text graph structure. Additionally, an efficient graph convolution operation is incorporated to enhance the information transfer and aggregation process, thereby reducing computational overhead.

This paper contributes in three main aspects: 1) introducing an adaptive strategy for constructing text graph structures to enhance accuracy; 2) designing a streamlined graph convolution operation to mitigate computational complexity; 3) The effectiveness and feasibility of the proposed method in actual text classification tasks are verified by a large number of experiments. It is hoped that the research in this paper can provide new ideas and technical support for text classification based on graph neural network, and promote the development of NLP field.

## II. RELATED WORK

As artificial intelligence continues to evolve at a rapid pace, a growing number of researchers and academics are delving into deep learning technologies. These technologies have found widespread application across various domains, including medical diagnosis [9-11], image classification[12-14], financial risk management [15-16], natural language processing (NLP), speech recognition, and text classification. Within the scope of NLP, text classification emerges as a pivotal research area, encompassing a wide array of methods and techniques. This section provides a comprehensive overview of traditional text classification approaches, underscores recent advancements in deep learning methodologies, and traces the development of Graph Neural Network-based methods. Furthermore, it critically examines their respective advantages, limitations, and the trajectories of their historical evolution.

Early methods of text classification relied heavily on statistical and machine learning models. Bag-of-Words model [17] and TF-IDF [18] are the most commonly used text representation methods. These methods transform text into word frequency vectors or TF-IDF vectors, ignoring the order and semantic relationships between words. Subsequently, traditional machine learning algorithms such as SVMs. While these methods perform well when dealing with small-scale data, they fall significantly short when it comes to capturing the deep semantics of text and processing large-scale data.

With the development of deep learning, CNNs and RNNs have been introduced into text classification tasks. CNN captures the local features of text, which is suitable for short text classification. However, CNNS are limited in their effectiveness when dealing with long text and long-distance dependencies. RNNS and their variants, such as Long Short-Term Memory [19] and Gated Recurrent Unit [20], effectively capture sequence information and context through cyclic structures. Significant progress has been made in text classification. However, these methods still have limitations when dealing with non-Euclidean structured data.

Graph Neural Networks have excelled in recent years in handling non-Euclidean data. GNN realizes information transfer and aggregation between nodes and their neighbors through graph convolution operation, which is suitable for processing text data with complex relational structure. The existing GNN-based text classification methods usually include the following steps: First, the construction of the text graph involves both word co-occurrence and syntactic dependency relationships; Secondly, models such as Graph Convolutional Networks and Graph Attention Networks are used to learn and classify node features[21].

Although GNN-based text classification methods show unique advantages in capturing semantic relationships and global structure information of text, there are still several challenges. First of all, the construction process of text graph structure is complicated and often needs to rely on external knowledge or pre-trained model, which leads to high computational cost. Secondly, the training and inference process of GNN model requires high resources, especially when dealing with large-scale text data, and there are bottlenecks in training efficiency and model performance.

The focus of this paper is a text classification optimization algorithm grounded in graph neural networks. It encompasses an adaptive strategy for graph construction and an efficient approach to graph convolution operations. The adaptive graph construction strategy dynamically adjusts the relationship between nodes and edges according to the text content, which improves the accuracy and construction efficiency of the graph structure. Based on the experimental findings, the proposed approach demonstrates clear superiority over traditional methods and existing GNN models across various public datasets. These results substantiate the method's effectiveness and advantages in enhancing text classification tasks.

In conclusion, this paper introduces a novel optimization algorithm that integrates traditional text classification methods, deep learning techniques, and graph-based neural network approaches. This proposal offers innovative ideas and technical foundations to enhance the precision and efficiency of text classification tasks.

III. THEORETICAL BASIS

*A. Graph neural networks*

The spectral domain-based graph neural network is a technique used for feature extraction and representation learning by leveraging spectral information inherent in graphs. It analyzes the eigenvalues and eigenvectors of the Laplacian matrix of the graph to comprehend its structure and features, enabling tasks such as node classification and graph classification to be performed effectively. This approach provides insights into the underlying characteristics of the graph, facilitating accurate classification tasks based on spectral properties.

First with regard to the representation of graphs and the Laplacian matrix, we assume that there is a $G = (V, E)$ where $V$ is nodes and $E$ is edges. Figure $G$ can be represented as the adjacency matrix $A$, where $A_{ij} = 1$ indicates that there is an $i$ and $j$, otherwise $A_{ij} = 0$. The Laplace matrix $L$ can be defined as $L = D - A$.

For Laplacian matrix $L$, its eigenvalues and eigenvectors can be obtained by spectral decomposition. Let $\lambda_1 \leq \lambda_2 \leq \cdots \leq \lambda_n$ be the eigenvalue of $L$, and the corresponding eigenvectors are $u_1, u_2, \ldots, u_n$. These eigenvectors form the spectral space of the graph $G$.

Graph convolution operations based on spectral domains can be expressed in the following form:

$$H^{(l+1)} = \sigma\left(\tilde{D}^{-\frac{1}{2}} \tilde{A} \tilde{D}^{-\frac{1}{2}} H^{(l)} W^{(l)}\right) \quad (1)$$

$H^{(l)}$ is the nodal eigenmatrix of the $l$ layer. $\tilde{A} = A + I$ is the result of adding A self-join to the adjacency matrix $A$.

Interlayer propagation of Graph Convolutional Networks (GCN) Interlayer propagation of GCN can be expressed in the following form:

$$H^{(l+1)} = \sigma(\hat{D}^{-\frac{1}{2}} \hat{A} \hat{D}^{-\frac{1}{2}} H^{(l)} W^{(l)}) \quad (2)$$

Among them: $\hat{A} = D^{-\frac{1}{2}} A D^{-\frac{1}{2}}$ adjacency matrix is symmetric normalization. $\hat{D}$ is the degree matrix of $\hat{A}$. GCNs can be trained for graph node classification tasks:

$$L = -\sum_{i=1}^{N}\sum_{k=1}^{K} y_{ik} \log(\hat{y}_{ik}) \quad (3)$$

Where: $y_{ik}$ is the label of whether the node $i$ belongs to the class $k$. $\widehat{y_{ik}}$ is the probability that the model predicts that the node $i$ belongs to the class $k$. By optimizing the loss function, the weight parameters in the graph neural network can be learned to complete the task of node classification.

While this is an overview of spectral domain based graph neural networks, recent research efforts have turned to exploring alternative matrix structures to optimize the performance of graph convolutional network (GCN) models. Among them, the model proposed by Mei et al. [22]

effectively captures and models the underlying structural information that is not explicitly expressed in the graph by introducing a learning distance function and a residual graph adjacency matrix. At the same time, double graph convolutional Network (DGCN) is a new approach, and a unique double graph convolutional architecture is proposed. DGCN consists of two sets of parallel graph convolution layers with shared parameters, using a normalized adjacency matrix and a matrix based on positive point mutual information. The positive point mutual information matrix captures the co-occurrence information between nodes by means of random walk. DGCN is also unique in that it synthesizes the output of the two-graph convolution layers to cleverly encode local and global structure information, thus reducing the dependence on the stacking of the multi-layer graph convolution layers.

Although spectral domain graph neural networks have a solid theoretical foundation and show good performance in practical tasks, they also expose several significant limitations. Firstly, many spectral domain graph neural network methods need to decompose Laplacian matrix to obtain eigenvalues and eigenvectors in the implementation process, which often brings high computational complexity. Although ChebNet and GCN simplify this step to some extent, the entire graph is still required to be stored in memory during the calculation process, which undoubtedly consumes a lot of memory resources. Secondly, the convolution operation of spectral domain graph neural network is usually carried out on the eigenvalue matrix of Laplacian matrix, which means that its convolution kernel parameters are not easily transferred when facing different graphs. Therefore, spectral domain neural networks are often limited to processing a single graph, which restricts their cross-graph learning ability and generalization ability, resulting in relatively few subsequent studies compared with spatial domain based graph neural networks.

The image neural network based on spatial domain method draws on the idea of traditional convolutional neural network (CNN) in image processing, extends the concept of convolution to the graph data structure, and defines the graph convolution operation according to the spatial correlation of nodes in the graph. As the pixels in Figure 1 constitute a two-dimensional grid structure, which can be regarded as a special form of topology diagram (see the left side of Figure 1), each pixel is regarded as a node, and adjacent pixels are connected to each other through edges. Similarly, when we apply the 3×3 convolution window on the image, the spatial-space-based graph convolution also simulates a similar process on the graph data. It integrates the feature sets of the central node and its neighboring nodes through convolution. This process is graphically shown on the right side of Figure 1. The core principle of spatial graph convolution lies in the propagation of node features and topological information along the edge structure of the graph, and the representation learning of the graph is carried out in this way, which is similar to the feature extraction and propagation of image data by CNN. In other words, spatial graph convolution can achieve the iterative updating and fusion of graph node features by simulating convolution behavior on graph data, thus playing a key role in the analysis and learning of graph data.

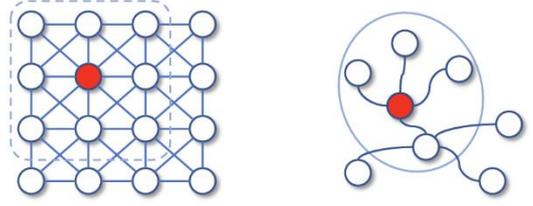

Fig. 1. Comparison of 2D convolution and graph convolution

Neural Network for Graphs (NN4G) is the first research achievement to implement spatial-domain based graph neural networks. It uses a composite neural structure with independent parameters to model the relationship between graphs, and extends the information. The graph convolution operation used by NN4G can be seen as a direct summation of the information of the neighbor nodes and the application of a residual network to hold the underlying information of the layer before the node. Therefore, the above process can be described by the following mathematical expression:

$$\mathbf{h}_v^{(k)} = f\left(\Theta^{(k)\bullet} \sum_{u \in \mathrm{N}(v)} \mathbf{h}_u^{(k-1)} + \sum_{i=1}^{|\mathrm{N}(v)|} \Theta^{(k)\bullet} \mathbf{A}_{vu}\mathbf{h}_u^{(k-1)}\right) \qquad (4)$$

Where $f(\cdot)$ as the activation function, $h_v^{(0)} = 0$. From the point of view of mathematical expression, the whole modeling process is the same as GCN. NN4G differs from GCN in that it uses an unnormalized adjacency matrix, which can result in very large differences in the scale of potential node information. The GraphSage[8-10] model was proposed to deal with the problem of large number of neighbors of nodes, and adopted the way of downsampling. In Figure 2, the model uses a series of aggregate functions for graph convolution to ensure that the output does not change as the node order changes. Among them, GraphSage uses three symmetric aggregation functions, namely mean aggregation, LSTM aggregation and pooling aggregation. The following is the mathematical expression used by the GraphSage model:

$$\mathbf{h}_v^{(k)} = \sigma\left(\mathbf{W}^{(k)} \cdot \mathrm{agg}_k\left(\mathbf{h}_v^{(k-1)}, \left\{\mathbf{h}_u^{(k-1)}, \forall u \in \mathrm{S}_N(v)\right\}\right)\right) \qquad (5)$$

Where $h_v^{(k)} = x_v$, $\mathrm{agg}_k(\cdot)$ is the aggregation function, $S_N(v)$ is $v$ neighbor nodes of a random sample. The proposal of GraphSage has brought positive significance to the development of graph neural networks. Inductive learning makes it easier to generalize graph neural networks, while neighbor sampling leads the trend of large-scale graph learning.

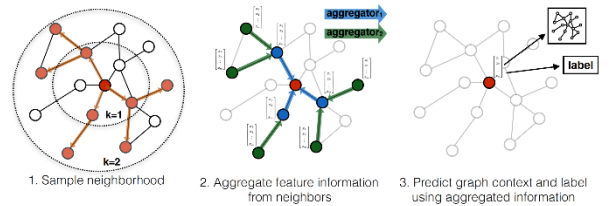

Fig. 2. GraphSage sampling and aggregation diagram

## B. Classification system

The development of classification system has experienced many important stages in the field of graphics, from traditional image processing methods to the rise of deep learning technology, and has made great progress and breakthroughs. In the early days, classification systems relied heavily on traditional image processing techniques. These methods are often based on hand-designed feature extractors and classifiers, such as edge detection, color histograms, texture features, etc. In the field of graphics, these techniques are widely used in character classification, object detection and other tasks. However, the performance of these methods is limited by the quality of the feature design and the complexity of the model. With the development of machine learning technology, support vector machine (SVM) and other methods are gradually introduced into the classification system. These methods can better deal with high dimensional data and complex classification tasks, and achieve good results. In the field of graphics, support vector machine and other methods have been applied to face classification, handwritten digit classification and other tasks, and have achieved some success.

The rise of deep learning technology marks a major advance in classification systems. In particular, the models has made a revolutionary breakthrough in the field of graphics classification systems. The CNN model realizes the feature learning and extraction of images through multi-layer convolution and pooling operations, thus achieving amazing performance in image classification, target detection, image segmentation and other tasks. The application scope of classification system in the field of graphics continues to expand, involving more complex scenes and tasks.

With the wide application of multimodal data, classification system begins to involve many types of data, such as image, text, speech and so on. Multimodal classification system makes use of the correlation between different types of data, and realizes the comprehensive analysis and classification of multimodal data by means of joint learning. In the field of graphics, multimodal classification system has been applied to image description generation, visual question answering and other tasks, providing people with a more rich and intelligent interaction way. To sum up, the development of classification system has experienced the evolution from traditional methods to deep learning technology in the field of graphics, and has made significant progress and breakthroughs. With the continuous development and innovation of technology, the application prospect of classification system in the field of graphics will be broader.

The following is a detailed explanation of the multimodal graphic classification system, which refers to the use of multiple types of data (such as images, text, speech, etc.) for comprehensive analysis and classification system. These systems can obtain information from data of different modes and achieve more accurate and comprehensive classification results by means of joint learning.

Data in multimodal graph classification systems are usually represented in the form of tensors. As shown in Figure 3 (taking face image data classification as an example), the system's work flow for image data classification includes steps such as data preprocessing, feature extraction, mode fusion and classification. First of all, the image data needs to be pre-processed, including scaling, cropping, normalization and other operations to ensure its quality and consistency. Then, deep learning models such as convolutional neural network (CNN) are used to extract features from the images and convert them into high-dimensional feature vectors. Then, the feature representation of the image is fused with other modal data, which can be serial, parallel or deep fusion. Finally, the fused feature representation is input into the classifier for classification, and the category or label of the image is predicted. After classification is complete, some post-processing operations may be required to improve the accuracy and stability of the results, including probabilistic calibration, result fusion, error correction, and so on. Through the combination of these steps, the multimodal graphic classification system can achieve accurate and comprehensive classification of image data. For text data, the Word Embedding representation is usually used, converting the text to a fixed-dimensional vector. For speech data, a Spectrogram is usually used to convert the sound waveform into a two-dimensional image.

The key of multimodal graph classification system is to realize the fusion of different modal data. Common fusion methods include serial fusion, parallel fusion and deep fusion. Serial fusion inputs different modal data into different models for processing, and then fuses the output of each model. Parallel fusion inputs different modal data into the same model for processing, and then fuses the output of the model. Deep fusion is to input different modal data into neural networks of different levels for processing, and then fuse the features of each level. Multimodal graph classification system usually adopts the way of joint learning to realize the comprehensive analysis and classification of various types of data. Joint learning can realize information exchange and joint training between different modal data by sharing partial parameters or alternating optimization. Common joint learning methods include multi-task learning, alternate training and deep joint learning.

In multimodal graph classification systems, the loss function typically comprises two main components: the mode-specific loss function and the cross-modal loss function. The mode-specific loss function evaluates the predictive performance of individual modal data, while the cross-modal loss function assesses the alignment and relationship between different modal data types. Commonly used cross-modal loss functions include correlation loss and contrast loss, among others.

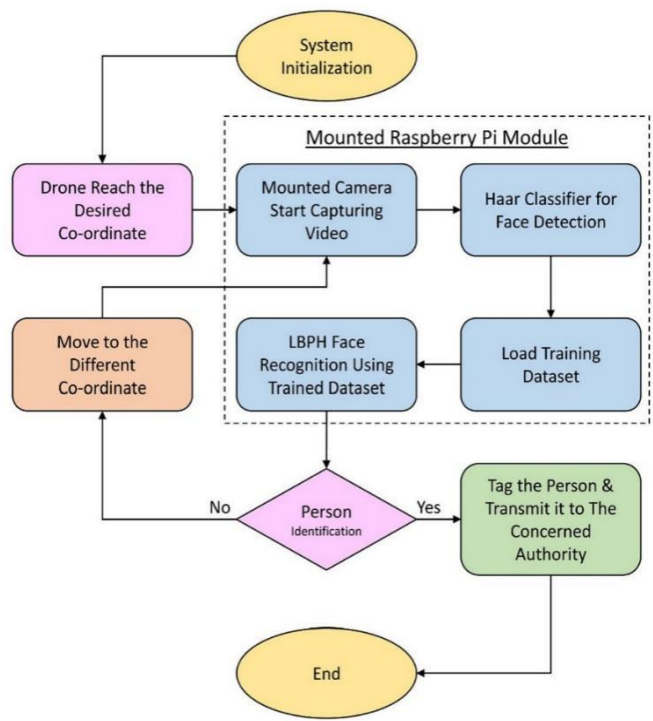

Fig. 3. Image data classification process.

## IV. TEXT CLASSIFICATION MODEL BASED ON GNN-MULTIMODAL INFORMATION

This paper combines Graph Neural Networks (GNNs) and Multi-Modal Information. A novel text classification model, GNN-MultiModal TextClassifier (GNN-MMC), is proposed. The objective of the model is to enhance both the accuracy and robustness of text classification, not only utilizing text data, but also integrating other modal information (such as images, audio, metadata, etc.) to comprehensively understand and classify text content.

The model architecture begins by transforming text into a graph structure, where nodes denote fundamental text elements (such as words, phrases, and sentences), and edges signify the connections among these elements (such as word co-occurrence and grammatical dependencies). Through the construction of the graph structure, the model is adept at capturing intricate semantic relationships and contextual information within the text. Specifically, given a text, a graph ($G = (V, E)$) is constructed from the co-occurrence of words or syntactic dependencies. Each node ($v \in V$) is initialized as a word embedding vector ($h_v^{(0)}$).

In addition to text data, the model also introduces information from other modes. For example, in social media analytics, text may be accompanied by images, audio, and user metadata. The information of these modes is transformed into corresponding feature vectors and integrated with the text graph. Suppose there is ($N$) mode information, and the eigenvector of each mode is expressed as ($m_i$), where ($i = 1, \ldots, N$). These modal features are transformed by a specific transformation function ($\phi_i(\cdot)$) into vectors ($m_i' = \phi_i(m_i)$) of the same dimension as the node features of the text graph.

In the graph neural network (GNN) layer, node features are updated by graph convolution operations. GNN learns the global semantic information by gradually updating the node representation through information aggregation of neighbor nodes. Specifically, the model uses Graph Attention Networks (GATs) architecture. In each layer of graph convolution, the feature update formula of node ($v$) is:

$$h_v^{(k+1)} = \sigma\left(\sum_{u \in N(v)} \alpha_{vu}^{(k)} \mathbf{W}^{(k)} h_u^{(k)}\right) \quad (6)$$

Where ($\mathcal{N}(v)$) represents the set of neighbor nodes of the node ($v$), ($W^{(k)}$) is the weight matrix of the ($k$) layer, ($\sigma$) is the activation function, ($\alpha_{vu}^{(k)}$) is the attention weight, calculated by the similarity between nodes:

$$\alpha_{vu}^{(k)} = \frac{\exp\left(\text{LeakyReLU}\left(\mathbf{a}^{(k)\cdot}[\mathbf{W}^{(k)} h_v^{(k)} \square \mathbf{W}^{(k)} h_u^{(k)}]\right)\right)}{\sum_{j \in N(v)} \exp\left(\text{LeakyReLU}\left(\mathbf{a}^{(k)\cdot}[\mathbf{W}^{(k)} h_v^{(k)} \square \mathbf{W}^{(k)} h_j^{(k)}]\right)\right)} \quad (7)$$

In order to integrate the information of different modes effectively, the model adopts the feature fusion strategy. Common methods include feature splicing, weighted fusion, and attention mechanisms. The multi-modal characteristics after fusion are expressed as follows:

$$\mathbf{h}_v^{\text{fused}} = \text{Concat}(\mathbf{h}_v^{(K)}, \mathbf{m}_1', \mathbf{m}_2', \ldots, \mathbf{m}_N') \quad (8)$$

Where ($h_v^{(K)}$) is the node feature updated by the convolution of the ($K$) layer graph. The fused multimodal features are input into subsequent classifiers, and finally, the fused features are input into the fully connected layer or other classifiers (such as SVM, Softmax classifier), and the category label of the output text is:

$$\hat{y} = \text{Softmax}(\mathbf{W}^{\text{cls}} \mathbf{h}_v^{\text{fused}} + \mathbf{b}^{\text{cls}}) \quad (9)$$

GNN-MMC offers several advantages. First of all, it can comprehensively process multi-modal information, which improves the comprehensiveness and accuracy of text classification. By integrating information such as images, audio, and metadata, models can capture important features beyond text. Secondly, through the graph neural network, the model can effectively capture the complex semantic relationships and global structure information in the text, which makes GNN-MMC have a significant advantage in dealing with long text and text with complex relationships. Finally, the graph neural network layer and multi-modal feature fusion layer of the model can learn and adjust the feature representation adaptively, which reduces the dependence on artificial feature engineering.

In summary, GNN-MMC provides an efficient and accurate text classification solution by combining graph neural networks and multimodal information. Based on the experimental outcomes, the proposed model exhibits significant superiority over traditional text classification methods and existing multimodal classification models across multiple datasets. In the future, GNN-MMC can be further expanded and optimized to deal with more diverse and complex practical application scenarios.

## V. EXPERIMENTAL ANALYSIS

### A. Data set

To assess and validate the performance of the GNN-MultiModal TextClassifier (GNN-MMC) model, this study opted to utilize the Twitter Multimodal Sentiment Analysis (TMMSA) dataset as the experimental dataset. The TMMSA dataset is a widely used standard dataset for multimodal text classification tasks and is particularly suitable for studying social media sentiment analysis. The TMMSA dataset contains a large number of emotionally tagged Twitter posts, each containing not only text content, but also associated images and other metadata, such as user information and post timestamps. This multimodal nature makes datasets particularly useful for models studying the integration of text and non-text information (such as images, metadata). The GNN-MMC model is designed to deal with this kind of complex information.

In the experiment, the TMMSA data set is preprocessed in detail. The dataset is managed using the Linked Data methodology [23], which consolidates various essential formats for academic research. This structured approach facilitates the interconnection of data across multiple datasets, thereby enhancing interoperability. This capability is particularly advantageous in fields such as machine learning and artificial intelligence, where the quality of data plays a critical role in training models and obtaining precise results. The text data goes through a standard text cleaning process, including the removal of stops, punctuation, stem extraction and word vectorization. The image data is preprocessed, including scaling, normalization, and image feature extraction using convolutional neural networks (e.g. ResNet, VGG). In addition, metadata such as user information and time stamps are transformed into appropriate numerical features for integration with text and image information. In experiments, the GNN-MMC model works by representing text as a graph structure, where nodes represent the basic units of text (such as words, phrases, sentences), and edges represent the semantic relationships between these units. The model uses graph neural network (GNN) for information aggregation and learning to capture global semantic information and complex contextual relationships. By combining multimodal information from images and metadata, GNN-MMC is able to understand and utilize information more comprehensively in text classification tasks, thereby improving classification accuracy and generalization.

Data preprocessing is a crucial step in any machine learning project, ensuring that the data is clean and valid before it is fed into the model. In this paper, we carried out a detailed data preprocessing process for the TMMSA data set used to prepare the data for use by the GNN-MultiModal TextClassifier (GNN-MMC) model. The following are the specific steps of data preprocessing:

a. Remove invalid characters and punctuation marks

First, remove invalid characters and punctuation from the text that could interfere with model training, such as special symbols, HTML tags, and so on. This step can easily be done with regular expressions or pre-processing libraries such as NLTK or Spacy.

b. Participle

Dividing text into sequences of words or phrases is a basic step in text processing. Word segmentation can be done using an off-the-shelf word divider (such as NLTK's word_tokenize), ensuring that the text is split into meaningful units.

c. Remove the stop word

Stop words are words that appear frequently in text analysis but usually do not contain useful information, such as "and", "the", etc. In the preprocessing process, removing the stops can reduce the noise.

d. Stem extraction or morphology reduction:

Commonly used techniques such as stemming or lemmatization are employed to simplify vocabulary complexity and reduce the feature space. They can convert words to their basic form, for example, converting both "running" and "ran" to "run."

e. Text vectorization

To convert processed text into numerical form, each word is mapped as a vector, usually using a Bag of Words (BoW) model or a word embedding model (e.g. Word2Vec, GloVe). These vectorized representations transform text information into a form that a machine learning model can process.

*B. Evaluation indicators*

One of the elements of our study is the evaluation of the classification performance which is based on the confusion matrix[24]. This matrix is used as a fundamental tool for predicting information into different categories. It organizes instances in four main quadrants which are: True Positives, True Negatives, False Positives, and False Negatives. In this case, TP indicates the instances of the right predicted positive cases, whilst TN means the count of the actual negative cases correctly labeled as negative in the dataset. On the other-hand FP is for those instances where the model by fault classifies negatives as positives and FN stands for the positive instances that have been incorrectly characterized as negatives.

The form of summing up some data, otherwise named a confusion matrix, is a set of ways to comprehend a model's precision level in various sections. A deep insight into these metrics (TP, TN, FP, and FN) allows us to have a complete knowledge concerning the program's predictive capabilities. It gives the possibility of measuring not only general exactness but also individual dimensions of the classifier's performance, consequently enabling a complete evaluation. The corresponding category Table 1 is as follows:

TABLE I. CONFUSION MATRIX OF SAMPLES

|  | positive sample | negative sample |
|---|---|---|
| positive | TP | FN |
| negative | FP | TN |

Accuracy is a major factor considered when assessing a model's performance. Together with F1 Score, it enables a deeper analysis of the precision and recall of the model. This widely used classification metric represents the ratio of correct predictions that are predicted correctly with the real outcomes regarding every single data point.

$$\text{Accuracy} = \frac{TP + TN}{TP + TN + FP + FN} \quad (10)$$

Conversely, the formula is such that, assess the model's capacity for categorization tasks, are both algorithms of the common practice. Classification accuracy indicates the percentage of correct predictions made.

As an explanation that is still well-structured, the term being specified is the best explanation of expected accuracy in the assessment of a model. Model performance evaluation usually implies the use of multiple metrics, such as Accuracy and F1 Score, which help in determining the effectiveness of a model. Here, the use of Accuracy predominates since its core idea is the percentage of correct forecasts out of the total samples. Basically, its computation is made using a specific formula.

The F1 score is created by the mean weighted of Precision and Recall, which brings both accuracy and recall into consideration. F1 scores are computed by the next formula:

$$F1 = \frac{2 \times \text{Precision} \times \text{Recall}}{\text{Precision} + \text{Recall}} \quad (11)$$

The Precision in model evaluation reads as the proportion of actual positive predictions among all the positive instances identified by the model; the Recall on the other hand is a distance function that provides an effective measure of the proportion of the real positive items that are properly designated as positive by the classification. Instead of Accuracy, which just focuses on the number of correct predictions, the F1 Score establishes a balance between Precision and Recall, consequently providing a more complex view of the model performance for different classes. The focus on both aspects makes it especially useful when evaluating the classification accuracy, especially in cases where the proper distinction between real positives and negatives that are not mistakenly labeled as positives is critical.

## C. Experimental setup

In the experimental part of this paper, we design the specific parameter Settings of the GNN-MultiModal TextClassifier (GNN-MMC) model in detail, and the necessary equipment requirements to ensure the repeatability and effectiveness of the experiment. First, we chose Graph Attention Networks (GATs) as the main graph neural network (GNN) layer structure. Compared with traditional GCN, GATs has better ability of neighborhood information aggregation and is suitable for processing complex unstructured data such as text. In the configuration of GAT, we set up 2 layers of GAT with 128 and 64 hidden units per layer and LeakyReLU as the activation function. These Settings are designed to take full advantage of the power of multi-layer networks to effectively capture complex semantic relationships and contextual information in text data. For multi-modal information fusion, we adopt a simple feature splicing strategy. Specifically, we spliced preprocessed text feature vectors, image features (2048-dimensional feature vectors extracted through ResNet), and metadata features (such as numerical representations of user information) directly together. This simple and direct fusion method helps to preserve the original characteristics of each mode information, while reducing the complexity and computational cost.

Additionally, a learning rate decay strategy was implemented to gradually adjust the learning rate throughout the training process, enhancing the model's convergence. The Batch Size of 64 was selected to strike a balance between available computational resources and model complexity. This ensures efficient processing of data batches in each iteration. For device requirements, we recommend using an NVIDIA GPU that supports CUDA acceleration for model training. Especially when dealing with large data sets, Gpus can significantly improve computational efficiency and speed up model training and inference processes. It is recommended to use a workstation or cloud server with sufficient memory to meet the large memory requirements that the GNN-MMC model may produce. In summary, with the above detailed parameter Settings and equipment requirements, we were able to fully evaluate and verify the performance of the GNN-MMC model on the TMMSA dataset. These Settings not only help ensure the reliability and validity of the experiment, but also provide a solid foundation and guidance for us to further explore and optimize the model.

## D. Experimental result

According to the experimental results of different baseline models in Table 2, we can deeply analyze the performance of each model and its comparison: First, the GNN-MMC model shows the best performance, with an accuracy of 96.01% and an F1 score of 97.88%. This shows that the GNN-MMC can significantly improve the accuracy and overall performance of text classification under the design of combining graph neural network and multi-modal information fusion. By using graph structure to capture complex semantic relationships and context information in text, and effectively integrating multi-modal information such as images and metadata, GNN-MMC can understand and classify text content more comprehensively, which is significantly reflected in the experimental results.

As a conventional graph neural network model, the GCN achieves an accuracy of 94.11% and an F1 score of 92.58%. This underscores the effectiveness of the GCN in traditional graph-based tasks. Although GCN is good at capturing local structures and relationships in graph data, it is somewhat poor at integrating multimodal information and processing global semantics, and thus slightly inferior to GNN-MMC in multimodal text classification tasks.

Finally, the RNN model performed relatively poorly, with an accuracy of 93.68% and an F1 score of 91.37%. As a traditional sequence model, RNN is strong in feature extraction and classification of short texts, but its performance is inferior to that of graph neural network model in processing long texts and complex semantic relationships. Especially without considering the multimodal information, the performance of RNN model is limited by the limitation of its sequence modeling. In summary, by comparing the experimental results of different baseline models, we can clarify the advantages of GNN-MMC model in multimodal text classification tasks. It combines advanced graph neural network technology and effective multi-modal information fusion strategy to provide an effective solution for processing complex text data, and has significant performance advantages and potential application prospects.

TABLE II. EXPERIMENTAL RESULTS AT DIFFERENT BASELINES

| Model | Acc | F1 score |
| --- | --- | --- |
| GNN-MMC | 96.01 | 97.88 |
| GCN | 94.11 | 92.58 |
| RNN | 93.68 | 91.37 |

## E. Ablation experiment

According to the ablation experiment results in Table 3, we can compare and analyze the performance of the GNN-MMC model and its components (GNN and MMC), so as to deeply understand the contribution and influence of each part in the overall performance of the model.

First, the GNN-MMC model showed the highest accuracy and F1 scores, at 96.01% and 97.88%, respectively. The results show that when GNN and MMC are applied at the same time, the model achieves the best comprehensive results on text classification tasks. GNN-MMC uses graph structure to capture complex semantic relationships in text effectively, and improves comprehensive understanding and classification of text content by integrating multi-modal information such as images and metadata.

The GNN model achieves an accuracy of 91.81% and an F1 score of 93.09%. However, when compared to GNN-MMC, which integrates multimodal information, the standalone GNN shows slight inadequacy, highlighting the critical role of multimodal fusion in enhancing overall classification performance. GNN performs well in learning local structures, but has limitations in dealing with global semantics and multi-modal information integration.

Furthermore, the MMC model achieves an accuracy of 92.99% and an F1 score of 92.58%. While the MMC model excels in integrating multimodal information, it lacks the graph neural network's capability to learn complex semantic relationships. This limitation hinders its performance in comprehending and accurately classifying long texts, which is where the GNN-MMC model excels.

In summary, the ablation experiment clearly demonstrated the advantages of GNN-MMC model using graph neural network and multi-modal information fusion. Gnn-mmc achieves excellent performance in text classification tasks by effectively combining GNN's global semantic learning with MMC's multimodal information integration. This study provides insights into the role of the model's components in complex tasks and points the way to designing more efficient multimodal text classification models in the future.

TABLE III. ABLATION RESULTS

| Model | Acc | F1 score |
|---|---|---|
| GNN-MMC | 96.01 | 97.88 |
| GNN | 91.81 | 93.09 |
| MMC | 92.99 | 92.58 |

VI. CONCLUSION

Based on the research and experiment of graph neural network and multimodal information fusion in text classification, we draw the following conclusions: This paper proposes an innovative text classification model, GNN-Multimodal TextClassifier (GNN-MMC), which combines the advantages of graph neural networks (GNN) and MultiModal information fusion (MMC). By introducing graph structure into the model, we can capture complex semantic relationships and contextual information in text data, while integrating multi-modal information (such as images, audio, metadata) further enriches the understanding of text content. According to the experimental results, the GNN-MMC model demonstrates significantly superior performance compared to traditional methods on the TMMSA dataset, achieving an accuracy rate of 96.01% and an F1 score of 97.88%. Ablation experiments further validated the contribution of GNN and MMC to the model performance. The performance of GNN and MMC when used alone is lower than that of GNN-MMC model when used in combination, which shows that comprehensive use of graph neural network and multi-modal information fusion are effective strategies when dealing with complex text tasks. Overall, the research in this paper not only promotes the application of graph neural networks in text classification, but also explores the potential of multimodal information in improving text understanding and classification accuracy. Future work could further optimize the structure and algorithms of the model and explore broader and complex application scenarios, such as sentiment analysis, event detection, etc., with a view to providing more powerful and flexible multimodal text analysis tools.